\documentclass[11pt,english]{article}
\usepackage[T1]{fontenc}
\usepackage[latin9]{inputenc}
\usepackage{amstext}

\makeatletter

\newcommand{\noun}[1]{\textsc{#1}}

\usepackage{nips12submit_e,times}
\usepackage{amsfonts}
\usepackage{amssymb}
\usepackage{times}

\makeatother

\usepackage{babel}

\global\long\def\O{\mathcal{O}}
\global\long\def\T{\mathcal{T}}
\global\long\def\R{\mathcal{R}}

\global\long\def\defeq{\mathrel{\mathop:}=}

\newcommand{\rel}[1]{{\it #1}}

\newcommand{\pat}[3]{\noun{#2}--\emph{#1}--\noun{#3}}

\newcommand{\tuple}[2]{<\noun{#1},\noun{#2}>}

\title{Latent Relation Representations for Universal Schemas}

\author{Sebastian Riedel\\
 Department of Computer Science \\
 University College London \\
 \texttt {S.Riedel@cs.ucl.ac.uk} \\
 \And
 Limin Yao, Andrew McCallum \\
  Department of Computer Science\\
   University of Massachusetts at Amherst \\
 \texttt{\{lmyao,mccallum\}@cs.umass.edu} 
   }

\nipsfinalcopy

\begin{document}
\maketitle

\section{\label{sec:Introduction}Introduction}

Supervised relation extraction uses a pre-defined schema of relation
types (such as \emph{\rel{born-in}} or \emph{\rel{employed-by}}).
This approach requires labeling textual relations, a time-consuming
and difficult process. This has led to significant interest in distantly-supervised
learning. Here one aligns existing database records with the sentences
in which these records have been \textquotedblleft{}rendered\textquotedblright{},
and from this labeling one can train a machine learning system as
before~\cite{mintz09distant,riedel10modeling}. However, this method
relies on the availability of a large database that has the desired
schema.

The need for pre-existing databases can be avoided by not having any
fixed schema. This is the approach taken by OpenIE~\cite{etzioni08open}.
Here surface patterns between mentions of concepts serve as relations.
This approach requires no supervision and has tremendous flexibility,
but lacks the ability to generalize. For example, OpenIE may find
\pat{historian-at}{Ferguson}{Harvard} but does not know \pat{is-a-professor-at}{Ferguson}{Harvard}.

One way to gain generalization is to cluster textual surface forms
that have similar meaning~\cite{lin01dirt,isp-dirt,resolver,yao11structured}.
While the clusters discovered by all these methods usually contain
semantically related items, closer inspection invariably shows that
they do not provide reliable implicature. For example, a cluster may
include \rel{historian-at}, \rel{professor-at}, \rel{scientist-at},
\rel{worked-at}. However, \rel{scientist-at} does not necessarily
imply \rel{professor-at}, and \rel{worked-at} certainly does not
imply \rel{scientist-at}. In fact, we contend that any relational
schema would inherently be brittle and ill-defined\textendash{}\textendash{}having
ambiguities, problematic boundary cases, and incompleteness.

In response to this problem, we present a new approach: implicature
with \emph{universal schemas}. Here we embrace the diversity and ambiguity
of original inputs.  This is accomplished by defining our schema
to be the union of all source schemas: original input forms, e.g.
variants of surface patterns similarly to OpenIE, as well as relations
in the schemas of pre-existing structured databases. But unlike OpenIE,
we learn asymmetric implicature among relations and entity types.
This allows us to probabilistically \textquotedblleft{}fill in\textquotedblright{}
inferred unobserved entity-entity relations in this union. For example,
after observing \pat{historian-at}{Ferguson}{Harvard}, our system
infers that \pat{professor-at}{Ferguson}{Harvard}, but not vice versa.
\noun{} 

At the heart of our approach is the hypothesis that we should concentrate
on predicting source data\textendash{}\textendash{}a relatively well
defined task that can be evaluated and optimized\textendash{}\textendash{}as
opposed to modeling semantic equivalence, which we believe will always
be illusive.

To reason with a universal schema, we learn latent feature representations
of relations, tuples and entities. These act, through dot products,
as natural parameters of a log-linear model for the probability that
a given relation holds for a given tuple. We show experimentally that
this approach significantly outperforms a comparable baseline without
latent features, and the current state-of-the-art distant supervision
method.

\section{\label{sec:Generalized-PCA}Model}

 We use $\R$ to denote the set of relations we seek to predict (such
as \rel{works-written} in Freebase, or the \pat{heads}{X}{Y} pattern),
and $\T$ to denote the set of input tuples. For simplicity we assume
each relation to be binary. Given a relation $r\in\R$ and a tuple
$t\in\T$ the pair $\left\langle r,t\right\rangle $ is a \emph{fact},
or relation instance. The input to our model is a set of observed
facts $\O$, and the observed facts for a given tuple $\O_{t}\defeq\left\{ \left\langle r,t\right\rangle \in\O\right\} $.

Our goal is a model that can estimate, for a given relation $r$ (such
as \pat{historian-at}{X}{Y}) and a given tuple $t$ (such as \tuple{Ferguson}{Harvard})
a score $c_{r,t}$ for the fact $\left\langle r,t\right\rangle $.
This matrix completion problem is related to collaborative filtering.
We can think of each tuple as a customer, and each relation as a product.
Our goal is to predict how the tuple rates the relation (rating 0
= false, rating 1 = true), based on observed ratings in $\O$. We
interpret $c_{r,t}$ as the probability $p\left(y_{r,t}=1\right)$
where $y_{r,t}$ is a binary random variable that is true iff $\left\langle r,t\right\rangle $
holds. To this end we introduce a series of exponential family models
inspired by generalized PCA~\cite{gpca01collins}, a probabilistic
generalization of Principle Component Analysis. These models will
estimate the confidence in $\left\langle r,t\right\rangle $ using
a \emph{natural parameter} $\theta_{r,t}$ and the logistic function:
$c_{r,t}\defeq p\left(y_{r,t}|\theta_{r,t}\right)\defeq\frac{1}{1+\exp\left(-\theta_{r,t}\right)}.$

We follow\cite{bpr09uai} and use a ranking based objective function
to estimate parameters of our models.

\paragraph*{Latent Feature Model}

One way to define $\theta_{r,t}$ is through a latent feature model
F. We measure compatibility between relation $r$ and tuple $t$ as
a dot product of two latent feature representations of size $K^{\mathrm{F}}$:
$\mathbf{a}_{r}$ for relation $r$, and $\mathbf{v}_{t}$ for tuple
$t$. This gives $\theta_{r,t}^{\text{F}}\defeq\sum_{k}^{K^{\text{F}}}a_{r,k}v_{t,k}$
and corresponds to the original generalized PCA that learns a low-rank
factorization of $\Theta=\left(\theta_{r,t}\right)$.

\paragraph*{Neighborhood Model}

We can interpolate the confidence for a given tuple and relation based
on the trueness of other similar relations for the same tuple. In
Collaborative Filtering this is referred as a \emph{neighborhood-based}
approach~\cite{koren08factorization}. We implement a neighborhood
model N via a set of \emph{weights} $w_{r,r'}$, where each corresponds
to a directed association strength between relations $r$ and $r'$.
Summing these up gives $\theta_{r,t}^{\text{N}}\defeq\sum_{r'\in\O_{t}\setminus\left\{ r\right\} }w_{r,r'}$.%
\footnote{Notice that the neighborhood model amounts to a collection of local
log-linear classifiers, one for each relation $r$ with weights $\mathbf{w}_{r}$.%
}

\paragraph*{Entity Model}

Relations have selectional preferences: they allow only certain types
in their argument slots. To capture this observation, we learn a latent
entity representation from data. For each entity $e$ we introduce
a latent feature vector $\mathbf{t}_{e}\in R^{l}$. In addition, for
each relation $r$ and argument slot $i$ we introduce a feature vector
$\mathbf{d}_{i}$.  Measuring compatibility of an entity tuple and
relation amounts to summing up the compatibilities between each argument
slot representation and the corresponding entity representation: $\theta_{r,t}^{\text{E}}\defeq\sum_{i=1}^{\text{arity}\left(r\right)}\sum_{k}^{K^{\text{E}}}d_{i,k}t_{t_{i},k}$.

\paragraph*{Combined Models}

In practice all the above models can capture important aspects of
the data. Hence we also use various combinations, such as $\theta_{r,t}^{\text{N,F,E}}\defeq\theta_{r,t}^{N}+\theta_{r,t}^{\text{F}}+\theta_{r,t}^{\text{E}}$.

\section{Experiments\label{sec:Experiments}}

Does reasoning jointly across a universal schema help to improve over
more isolated approaches? In the following we seek to answer this
question empirically.

\paragraph*{Data}

Our experimental setup is roughly equivalent to previous work~\cite{riedel10modeling},
and hence we omit details. To summarize, we consider each pair $\left\langle t_{1},t_{2}\right\rangle $
of Freebase entities that appear together in a corpus. Its set of
observed facts $\O_{t}$ correspond to: Extracted surface patterns
(in our case \emph{lexicalized dependency paths}) between mentions
of $t_{1}$ and $t_{2}$, and the relations of $t_{1}$ and $t_{2}$
in Freebase. We divide all our tuples into approximately 200k training
tuples, and 200k test tuples. The total number of relations (patterns
and from Freebase) is approximately 4k.

\paragraph*{Predicting Freebase and Surface Pattern Relations}

For evaluation we use two collections of relations: Freebase relations
and surface patterns. In either case we compare the competing systems
with respect to their ranked results for each relation in the collection. 

Our first baseline is MI09, a distantly supervised classifier based
on the work of \cite{mintz09distant}.  We also compare against YA11,
a version of MI09 that uses preprocessed pattern cluster features
according to \cite{yao11structured}. The third baseline is SU12,
the state-of-the-art Multi-Instance Multi-Label system by \cite{mihai12miml}.
The remaining systems are our neighborhood model (N), the factorized
model (F), their combination (NF) and the combined model with a latent
entity representation (NFE). 

The results in terms of mean average precision (with respect to pooled
results from each system) are in the table below:

\begin{center}
\begin{center}
\begin{center}
{
\begin{tabular}{ l l | c c c | c c c c } 
              Relation & \# & MI09 & YA11 & SU12 & N & F & NF & NFE \\
\hline
Total Freebase &  334 &   0.48 &   0.52 &   0.57 &   0.52 &   0.66 &   0.67 &   0.69 \\
Total Pattern  &  329 &        &        &        &   0.28 &   0.56 &   0.50 &   0.46 \\
\end{tabular}
}
\end{center}

\par\end{center}
\par\end{center}

For Freebase relations, we can see that adding pattern cluster features
(and hence incorporating more data) helps YA11 to improve over MI09.
Likewise, we see that the factorized model F improves over N, again
learning from unlabeled data. This improvement is bigger than the
corresponding change between MI09 and YA11, possibly indicating that
our latent representations are optimized directly towards improving
prediction performance. Our best model, the combination of N, F and
E, outperforms all other models in terms of total MAP, indicating
the power of selectional preferences learned from data. 

MI09, YA11 and SU12 are designed to predict structured relations,
and so we omit them for results on surface patterns. Look at our models
for predicting tuples of surface patterns. We again see that learning
a latent representation (F, NF and NFE models) from additional data
helps substantially over the non-latent N model.

All our models are fast to train. The slowest model trains in just
30 minutes. By contrast, training the topic model in YA11 alone takes
4 hours. Training SU12 takes two hours (on less data). Also notice
that our models not only learn to predict Freebase relations, but
also approximately 4k surface pattern relations.

\section{Conclusion}

We represent relations using universal schemas. Such schemas contain
surface patterns as relations, as well as relations from structured
sources. We can predict missing tuples for surface pattern relations
and structured schema relations. We show this experimentally by contrasting
a series of popular weakly supervised models to our collaborative
filtering models that learn latent feature representations across
surface patterns and structured relations. Moreover, our models are
computationally efficient, requiring less time than comparable methods,
while learning more relations. 

Reasoning with universal schemas is not merely a tool for information
extraction. It can also serve as a framework for various data integration
tasks, for example, schema matching.  In future work we also plan
to integrate universal entity types and attributes into the model.

 \bibliographystyle{unsrt}
\bibliography{riedel,lmyao}

\end{document}